%% file: main.tex
\documentclass[dvipsnames,format=sigconf,nonacm,review=false,timestamp=false]{acmart}
\usepackage{xcolor}
\usepackage{graphicx}

\usepackage[T1]{fontenc}
\usepackage[utf8]{inputenc}
\usepackage[linesnumbered, ruled, vlined]{algorithm2e}
\usepackage{multirow}

\usepackage{upquote}
\usepackage{svg}
\usepackage{caption}
\usepackage{subcaption}
\usepackage{xspace}
\usepackage{array}
\usepackage[section]{placeins}

\AtBeginDocument{%
  \providecommand\BibTeX{{%
    \normalfont B\kern-0.5em{\scshape i\kern-0.25em b}\kern-0.8em\TeX}}}

\newcommand{\algoname}{HOTGP\xspace}

\newcommand{\code}[1]{\texttt{#1}}
\begin{document}

\sloppy

\title{\algoname --- Higher-Order Typed Genetic Programming}


\author{Matheus Campos Fernandes}
\email{fernandes.matheus@ufabc.edu.br}
\orcid{0000-0002-5118-2399}
\affiliation{%
  \institution{Federal University of ABC (UFABC)}
  \city{Santo André}
  \state{São Paulo}
  \country{Brazil}
}
\author{Fabrício Olivetti de França}
\email{folivetti@ufabc.edu.br}
\orcid{0000-0002-2741-8736}
\affiliation{%
  \institution{Federal University of ABC (UFABC)}
  \city{Santo André}
  \state{São Paulo}
  \country{Brazil}
}
\author{Emilio Francesquini}
\email{e.francesquini@ufabc.edu.br}
\orcid{0000-0002-5374-2521}
\affiliation{%
  \institution{Federal University of ABC (UFABC)}
  \city{Santo André}
  \state{São Paulo}
  \country{Brazil}
}


\begin{abstract}
Program synthesis is the process of generating a computer program following a set of specifications, which can be a high-level description of the problem and/or a set of input-output examples. 
The synthesis can be modeled as a search problem in which the search space is the set of all the programs valid under a grammar. 
As the search space is vast, brute force is usually not viable and search heuristics, such as genetic programming, also have difficulty navigating it without any guidance. 
In this paper we present \algoname, a new genetic programming algorithm that synthesizes pure, typed, and functional programs. 
\algoname leverages the knowledge provided by the rich data-types associated with the specification and the built-in grammar to constrain the search space and improve the performance of the synthesis. 
The grammar is based on Haskell's standard base library (the synthesized code can be directly compiled using any standard Haskell compiler) and includes support for higher-order functions, $\lambda$-functions, and parametric polymorphism. 
Experimental results show that, when compared to $6$ state-of-the-art algorithms using a standard set of benchmarks, \algoname is competitive and capable of synthesizing the correct programs more frequently than any other of the evaluated algorithms.

\end{abstract}



\maketitle

\begin{acks}
This research was partially funded by Funda\c{c}\~{a}o de Amparo \`{a} Pesquisa do Estado de S\~{a}o Paulo (FAPESP), grant numbers \#2021/12706-1, \#2019/26702-8, \#2021/06867-2 and Coordena\c{c}\~{a}o de Aperfei\c{c}oamento de Pessoal de N\'{i}vel Superior - Brasil (CAPES).
\end{acks}

\section{Introduction}
\label{sec:introduction}
\input{sections/introduction}

\section{Related Work}
\label{sec:related_work}
\input{sections/related_work}


\section{Higher-Order Typed Genetic Programming}
\label{sec:methods}
\input{sections/methods}

\section{Experimental Results}
\label{sec:results}
\input{sections/results}

\section{Conclusion}
\label{sec:conclusion}
\input{sections/conclusion}


\bibliographystyle{ACM-Reference-Format}
\bibliography{references}

\end{document}

%% file: sections/introduction.tex
Program Synthesis (PS) is the task of creating a computer program, in algorithmic form, based on a set of specifications~\cite{gulwani2010dimensions}.
A program specification is a high-level description of the objective of the program~\cite{koza2005genetic}. This specification can have different formats, from natural language to a more formal notation. 
A common approach to specifying a program to solve a problem is providing a set of input-output examples, with special attention to edge cases, to reduce the set of ambiguous solutions. This is called Inductive Synthesis, or Programming-by-Examples (PBE)~\cite{gulwani2017program}.
In this case, the task of the synthesizer is to find a program that correctly maps each pair of input-output provided by the examples. Depending on the completeness of the provided examples, the specification might be ambiguous leading to many alternative programs that do not behave as intended, even if they correctly map those examples.

The generation of a program can be modeled as a search where the search space contains the set of all possible programs valid under a pre-specified grammar.
The objective is to find a program that meets the specification.
The search space size makes it impractical to employ a naive approach for selecting the best candidate from the enumeration of all possible programs.
For this reason, the PS is often done using a (meta-)heuristic approach, usually Genetic Programming (GP)~\cite{koza1992genetic}.

Even though this approach has presented some success in standard PS benchmarks~\cite{
helmuth2015detailed,helmuth2018umad,kelly2019improving,hemberg2019domain,helmuth2022problem}, it is still incapable of finding solutions to some tasks that are trivial for humans.
Some of the reasons for this are:

\begin{enumerate}
    \item Traversing the search space is challenging, as sometimes a small change in the program code significantly impacts its output;
    \item Without additional information about the program, the search relies on the completeness of the examples;
    \item Some synthesizers can create stateful programs that can have unpredictable behavior depending on how the states are changed.
\end{enumerate}

To illustrate these difficulties, let us take as an example a specification for a program that, given a list \code{x}, returns the sum of all the elements which are smaller than 100. 
A valid imperative style program is presented in \autoref{alg:example}.
Changes to any of the initial values of \code{s} (line~\ref{line:sum}), \code{i} (line~\ref{line:for}) or the constant $100$ (line~\ref{line:compare}) can drop the accuracy from $100\%$ to $0\%$. 
On the other hand, for instance, if the input-example lists do not contain values between $95$ and $100$, a constant such as 98 in line~\ref{line:compare} will be enough to achieve $100\%$ accuracy on the training set, even though the code will produce wrong results when one considers all possible inputs.
Finally, any additional statement inside the loop body that affects the value of either \code{i} or \code{s} might also decrease the achieved accuracy.

\begin{algorithm}[t]
\caption{One solution to the example specification.}\label{alg:example}
\SetKwFunction{solution}{solution}%
\SetKwProg{Fn}{\textbf{function}}{\string:}{}%
\SetKwFunction{length}{length}%

\DontPrintSemicolon

\Fn{\solution{x}}{
    \DataSty{s} $\gets 0$\; \label{line:sum}
    \For{$i \gets 0;\ i < \length{x};\ i \gets i + 1$  \label{line:for} }{
        \If{
        x[i] < 100  \label{line:compare}
        }{
            \DataSty{s} $\gets$ \DataSty{s} + \DataSty{x[i]}\;
        } 
    }
    \Return \DataSty{s}
}
\end{algorithm}

One possible solution to alleviate these problems is to employ a typed and purely functional paradigm.
In this paradigm, a program is a \emph{pure} function and is defined as the composition of pure functions.

A pure function has, by definition, the fundamental property of \emph{referential transparency}~\cite{sondergaard1990referential}. This means that any expression (including the whole program which is, by itself, an expression) can safely be substituted by the result of its evaluation.
This property makes input transformations explicit and predictable, constraining the search space to only functions with no side effects.
Finally, a typed language contains information about the input and output types which helps us to constrain the search space further.

Moreover, if we also allow for parametric polymorphism\footnote{Also known as generics in some programming languages.}, we can effectively constrain the search space to contain only well-formed programs.
Take, for example, the type signature \code{$\forall$~a.~a~$\to$~a}. This signature only allows a single implementation, which is the identity function.
Any other implementation would either violate referential transparency or the type information.
Albeit extreme, this example shows how the combination of these properties can constrain the search space, easing the task of the PS algorithm~\cite{montana1995strongly}.

While the mentioned features already allow for an expressive language, typical functional programming languages also provide constructs for the implementation of higher-order functions~\cite{jones1995system}.
In this context, a higher-order function is a function that receives a function as one of its arguments.
Commonly used higher-order functions are \code{map, filter, fold}, which generalize many common patterns required by a program.

This work proposes a new GP algorithm, named \algoname (Higher-Order Typed Genetic Programming), that searches for pure, typed, and functional programs. 
The grammar supports higher-order functions, parametric polymorphism in functions, and parametric types (such as lists and tuples).
\algoname was evaluated against $29$ benchmark problems and its results compared to $6$ other algorithms from the literature. 
Results show that a pure functional approach can significantly improve the results of the standard GP algorithm in terms of the frequency that it finds correct programs.

The remainder of this paper is organized as follows.
Section~\ref{sec:related_work} presents related work. 
Section~\ref{sec:methods} describes \algoname. 
The experimental evaluation is outlined in Section~\ref{sec:results}, and we conclude in Section~\ref{sec:conclusion}.

%% file: sections/related_work.tex
To the best of our knowledge, Automatic Design of Algorithms Through Evolution (ADATE)~\cite{olsson1995inductive} is the earliest example of PS targeting functional code.
This work aimed at synthesizing recursive ML language programs using incremental transformations. 
The algorithm starts with an initial program described by the token "\code{?}" that always returns a \emph{don't know}\footnote{This is equivalent to a function that always returns null.} value.
After that, ADATE systematically expands the expression into a pattern matching of the input type, synthesizing a program for each branch of the pattern match, and replacing the general case with a recursive call. 

\citet{montana1995strongly} proposes the Strongly Typed Genetic Programming (STGP) algorithm, an adaption of GP that considers the types of each function and terminal during the PS.
The purpose of taking types into consideration is to further constrain the search space by allowing only correctly-typed programs to exist (\emph{i.e.}, programs in which all functions operate on values with the appropriate data types).
In contrast to standard GP, where a given nonterminal must be capable of handling any data type, STGP imposes extra constraints to enforce type-correctness.
Another important contribution of the STGP is that the types of the nonterminals can employ parametric polymorphism.

The main benefit of having parametric polymorphism is that there is no need for multiple similar functions whose difference is only in their types.
Experiments on four different problems (regarding matrix and list manipulations) have shown that STGP generally outperforms untyped GP.

STGP and a standard untyped GP were compared by \citet{haynes1995strongly} using the ``Pursuit Problem''.
This problem models a game where four predators pursue a prey.
The goal is to create an algorithm for the predators to capture the prey as fast as possible. The prey always runs away from the nearest predator, and the predators only have information about themselves and the prey, but not about the other predators.
Results show that a good STGP program can be generated faster than a good GP program. Moreover, the best STGP program has a higher capture rate than the best GP program.

PolyGP~\cite{yu1997polygp,yu2001polymorphism} extends STGP with support to higher-order functions and $\lambda$-functions. 
It also differs from STGP by using a type unification algorithm instead of a lookup table to determine the concrete types when using polymorphic functions. 
The $\lambda$-functions use the same initialization procedure of the main PS, but the available terminals are limited to the input parameters.
Because these $\lambda$-functions do not have any type restriction, they can be invalid in which case it must be discarded and regenerated. 
The overall algorithm is a simple search for a composition of $\lambda$-functions with a user-defined set of terminals and nonterminals as in STGP.

\citet{katayama2005systematic} proposes MagicHaskell, a breadth-search approach that searches for a correctly-typed functional program using SKIBC~\cite{turner1979new} combinators. 
This simplifies the PS by reducing the search space. 
MagicHaskell also introduces the use of the de Bruijn lambda to find equivalent expressions and memoization to improve performance~\cite{de1972lambda}. 
Additionally, it implements fusion rules to simplify the synthesized program further. This particular approach was reported not to work well with larger problems~\cite{10.1145/3067695.3082533}.

Strongly Formed Genetic Programming (SFGP)~\cite{sfgp2012} is an extension to STGP. 
SFGP not only assigns known data-types to terminals but also node-types to functions. 
A node-type identifies if a given node is a variable, an expression, or an assignment.
Each subtree of a function can also be required to be of a certain node-type.
The authors argue that this extra information is helpful to build correctly typed \emph{imperative} programs (\emph{e.g.}, the first child of an assignment must have the ``Variable'' node-type and match the data-type of the second child). 
They conducted experiments on $3$ datasets, with a reduced grammar that deals mainly with integers, and reported high success rates with a lower computational effort than competing methods.

\citet{santos2020refined} discuss desiderata for PS approaches by further constraining the search space, similar to what is done by STGP.
They propose the use of Refinement Types to this aim.  As this
is an ongoing project, to the best of our knowledge, there are still
no experimental evaluations or comparative results.

\citet{pantridge2022functional} proposes an adaptation of the Code Building Genetic Programming (CBGP)~\cite{pantridge2020code} as a means to incorporate elements of functional programming such as higher-order functions and $\lambda$-functions.
CBGP uses the same representation of PushGP with three primary constructs: \code{APP}, to apply a function; \code{ABS}, to define a function of $0$ or more arguments and; \code{LET}, to introduce local variables in the current scope. 
It also uses concepts from type theory to ensure the correctness of the polymorphic types. 
CBGP achieved higher generalization rates for a subset of benchmark problems. 
However, for other problems, the generalization rate was close to $0$.
The authors noted that the evolutionary search avoided using $\lambda$-functions and preferred to employ pre-defined functions in higher-order functions such as \code{map}. 
These results show some indirect evidence of the benefits provided by type-safety to PS, in particular, with regard to the stability of the solutions over different executions of the search algorithm.

In this same line, \citet{garrow2022functional} compared the generation of Python and Haskell programs using a grammar-guided system~\cite{manrique2009grammar}.
Similar to our work, they employ a different grammar for each set of types instead of a different grammar per benchmark problem. 
Their approach supports higher-order functions, but limits the function arguments to pre-defined commonly used functions. 
Experimental results showed that the Haskell version consistently outperforms Python in most selected benchmarks. 
Implementing general $\lambda$-functions was left as future work by the authors since that would add complexity to the search space and must be carefully handled as a different construct from the main program.

\citet{he2022incorporating} investigate the reuse of already synthesized programs as subprograms to be incorporated in the nonterminal set.
The main idea is that, if the algorithm has already synthesized solutions to simpler tasks, these solutions can be used to build more complex solutions, in an incremental process. 
Their results show a significant benefit could be obtained by adding handcrafted modules in $4$ selected benchmarks.

\citet{g3p} criticize a common technique in GP, which is to provide a different grammar for each problem. 
They argue that this leads to difficulties in grammar reuse, as they are specifically tailored to each problem.
They propose a general grammar to the G3P algorithm and perform experiments on the benchmark introduced by \citet{helmuth2015detailed}.
Since the proposed grammar had difficulty with the benchmark problems involving characters and strings, the authors proposed an improved and expanded grammar leading to G3P+~\cite{g3pe}.

%% file: sections/methods.tex
This section introduces Higher-Order Typed Genetic Programming (\algoname). 
To the best of our knowledge, STGP was among the first to propose and employ types for GP. 
As such, it naturally has influenced following works, such as \citet{sfgp2012}, \citet{santos2020refined}, and \algoname.
We now present the main concepts needed for typed GP which are shared by all these synthesizers.

In Strongly-Typed Genetic Programming (SGTP), every terminal has an associated data type, and nonterminals have associated input types and one output type. 
To enforce correctness, the algorithm imposes two constraints: 
i) the root of the tree must have the same output type as the intended program output type; 
ii) every non-root node must have the output type expected by its parent.

Due to these restrictions, the main components of the evolutionary search must be adapted. 
At every step of the initialization process, a node will be considered only if it matches the type expected by its parent node.
STGP also builds type-possibility tables to keep track of which data-types can be generated by a tree of each depth, one for each initialization method (\emph{grow} and \emph{full}).
Those tables are dictionaries whose keys represent the depths and the values the types representable by trees of that depth, for \emph{full} or \emph{grow}. At depth $0$, they contain only terminals.
At depth $i$, they contain the output types of all the functions that take the types at $i-1$ as an argument; and for \emph{grow}, it also contains the types at $i-1$.

The mutation operator replaces a random subtree with a new subtree
generated with the same algorithm of the initialization procedure, using the \emph{grow} method.
The crossover, as expected, also takes into account the types. 
The crossover point of the first parent is
chosen entirely at random, while the point of the second parent is
limited to those whose type is the same as the first parent. 
If no such candidate exists, it returns one of the parents.
There are also some additional changes to the original GP algorithm
regarding the evolutionary process such as the use of \emph{steady-state replacement}~\cite{syswerda1991study} and \emph{exponential fitness normalization}, which select parents for reproduction based on their ranks. The probability of picking the $n^\mathit{th}$ best individual is given by $p(n) = P_\mathit{scalar} \times p(n-1)$, where $0 < P_\mathit{scalar} < 1$ is a hyperparameter.

\citet{montana1995strongly} also argues in favor of handling runtime errors as part of the evolutionary process, penalizing individuals that present them.
This is the opposite of the original GP approach~\cite{koza1992genetic}, which enforces that a value must always be returned.

STGP also introduces the \code{Void} type for functions that do not return anything (\emph{i.e.}, procedures) and local variables which can be statefully changed during computation.

In contrast to STGP, \algoname introduces higher-order functions and $\lambda$-functions, drops the support for impure functions, and uses a general-use grammar extracted from Haskell's base library. 
The main differences, detailed in the next subsections, are:

\begin{itemize}
    \item \algoname builds programs using a pure functional program paradigm (a subset of the Haskell programming language) while STGP is modeled after a combination of typed-LISP and ADA allowing impure functions (see Section~\ref{sec:grammar});
    \item Since \algoname is designed to only support pure functions, all side effects, including local variables (mutable state) and IO, are disallowed by design (see Sec.~\ref{sec:grammar});
    \item As we shifted to a different language, appropriate changes to the grammar were performed (see Section~\ref{sec:grammar});
    \item Instead of specifying a strict set of terminals and non-terminals which are specific to each problem, we specify generic sets based on the input and output types\footnote{This is a design choice that was not explored by STGP nor PolyGP.} (see Section~\ref{sec:grammar});
    \item Moreover, we use a more generic set of non-terminals (all available in the standard Haskell base library) instead of very specific functions that often need to be implemented by the user. 
    This characteristic, combined with the use of a subset of the Haskell language, allows for all the synthesized code to be immediately consumed by a Haskell compiler without modification (see Section~\ref{sec:grammar});
    \item Finally, \algoname has support for higher-order functions (functions that accept $\lambda$-functions as input) to handle advanced constructs in the synthesized programs (see Section~\ref{sec:lambdas});
\end{itemize}

\subsection{Functional Grammar}\label{sec:grammar}

Even though both \algoname and STGP share the use of strong types, in both experimental evaluations of STGP~\cite{montana1995strongly, haynes1995strongly}, the authors employed a limited grammar specifically crafted for each one of the benchmark problems. 
For example, to solve the Multidimensional Least Squares Regression problem, they used a minimal set of functions with matrix and vector operators such as \emph{matrix\_transpose, matrix\_inverse, mat\_vec\_mult, mat\_mat\_mult}. 
Instead, this paper uses a more general set of functions, common to all problems, all of which were extracted from the standard Haskell base library.

We argue that, in a practical scenario, providing only the functions needed for each problem is undesirable since it involves giving too much information to the algorithm. 
This is, in our opinion, not ideal since this piece of information might not be readily available beforehand. 
A much more reasonable demand on the user is to ask them for the acceptable result type for each problem.
This kind of information usually only requires as much intuition on the problem as providing examples.

\algoname primitive types currently includes $32$-bit integers, single-precision floating-point numbers, booleans, and UTF8 characters. The following parametric types are also supported: pairs (2-tuples); linked lists; and $\lambda$-functions.
Types can be combined to create more complex types, \emph{e.g.}, a list of pairs of $\lambda$-functions or, something simpler such as a string (represented as a list of characters).

As a consequence of using a subset of the Haskell language, \algoname precludes the use of impure functions. 
The use of pure functions is often associated to a reduction of the number of possible bugs~\cite{ray2014large}. 
An essential property of pure functions is that, being without side effects, they are easier to compose. 
Thus, whenever the return type of one function is the same as the input type of another function, they can be composed to form a new, more complex pure function. 

The full list of the functions allowed by \algoname's grammar is shown in \autoref{tab:functions}\footnote{For the sake of space and legibility, in this text we represent pairs and lists using ML-inspired conventions: \code{(7, 42)} is a pair containing \code{7} and \code{42}, and \code{[42, 7, 6]} is a list with elements \code{42}, \code{7} and \code{6}. Similarly, we write \code{[a]} in lieu of the type \code{List a} and \code{(a, b)} in lieu of the type \code{Pair a b}.}.
Most functions are common operations for their specific types. 
Since we employ a strongly-typed language, we also require conversion functions. 
Additional functions of common use include
sum and product for lists of numbers (integers and floating points),
\code{Range}, which generates a list of numbers (equivalent to Haskell's \code{[x,y..z]});
\code{Zip}, that pairs the elements of two lists given as input;
\code{Take}, that returns the first $n$ elements of a list;
and \code{Unlines}, that transforms a list of strings into a single string, joining them with a newline character.
In particular, \code{Unlines} is needed for the benchmarks requiring the program to print text to the standard output (in our case, since we are working on a pure language, we simply return the output string).

\input{tables/functions_smaller}

It is worth noting that we included three \emph{constructor} functions in the grammar: \code{ToPair}, \code{Cons}, and \code{Singleton}. This is a deliberate choice to simplify the grammar. 
Let us take 2-tuples (pairs) as an example. 
Our grammar must be able to cope with constructions such as \code{(1, 2)} or \code{(1 + 2, 3 * 4)} (pairs of literals and pairs of expressions). 
However these same pairs can be easily represented as applications of \code{ToPair}.
The first example can be represented as \code{ToPair 1 2}, which means applying the \code{ToPair} function to the arguments \code{1} and \code{2}.
Following the same representation, the second example becomes just \code{ToPair (AddInt 1 2) (MultInt 3 4)}.

In other words, the construction of a pair is a simple function application with no special treatment by our grammar. This has the added benefit of being directly compatible with the mutation and crossover operators already defined for regular nodes. Under the same reasoning, the evolution process can generate linked lists using a combination of the \code{Cons} and \code{Singleton} functions. For example, the list of the literals \code{1}, \code{2}, \code{3} can be represented as \code{Cons~1~(Cons 2 (Singleton 3))}; and the list of the expressions \code{1 + 2}, \code{3 * 4}, \code {5 - 6} can be represented as \code{Cons (AddInt 1 2) (Cons (MultInt 3 4) (Singleton (SubInt 5 6)))}.
As was also the case with pairs, this has the added benefit of enabling crossover and mutation to happen on just the head or just the tails of such lists.

\algoname also allows the user to select which types the program synthesis algorithm can use, to constrain the search space further.
Whenever the user selects a subset of the available types, the non-terminal set is inferred from \autoref{tab:functions} by selecting only those functions that support the selected types.
For example, if we select only the types \code{Int} and \code{Bool} we would allow functions such as \code{AddInt}, \code{And}, \code{GtInt}, but would not allow functions such as \code{Head}, \code{Floor}, \code{ShowInt}.

For a future implementation of this algorithm, we plan to add support to ad-hoc polymorphism, employing Haskell's type-classes, so we can simply have \code{Add, Mult, Sub} that determine their types by the context instead of having specific symbols for each type.

Another important distinction from STGP to \algoname is the absence of the \code{Void} type, and constructs for creating local variables.
Therefore, impure functions and mutable state are not representable by \algoname's grammar. 
By construction, \algoname does not allow side effects and can only represent pure programs.
On the other hand, similarly to STGP, runtime errors (such as divisions by zero) can still happen.
When they do, the fitness function assigns an infinitely bad fitness value to that solution.

\subsection{Higher-order Functions and \texorpdfstring{$\lambda$}{λ}-functions}\label{sec:lambdas}

The main novelty of \algoname is the use of higher-order functions. To that end, adding support to $\lambda$-functions is essential.
A $\lambda$-function, or anonymous function, or simply lambda, is a function definition not bound to a name.
As first-class values, they can be used as arguments to higher-order functions.

The introduction of lambdas requires additional care when creating or modifying a program.  
When evaluated, \algoname's lambdas only have access to their own inputs, and not to the main program's.
In other words, they do not capture the environment in which they were created or in which they are executed.
This means that lambda terminals can be essentially considered ``sub-programs'' inside our program, and are generated as such.
We use the same initialization process from the main programs, using
the function type required by the current node and employing the \emph{grow} method.
However, two additional constraints must be respected.

Constraint 1 requires all lambdas to use their argument in at least one of its subtrees, which significantly reduces the possibility of the creation of a lambda that just returns a constant value. 
We argue that, for higher-order-function purposes, a lambda is required to use its argument in order to produce interesting results; otherwise the program could be simplified eliminating the use of this higher-order function and returning a constant\footnote{This is only true because \algoname's grammar precludes the generation of expressions with side effects.}.

Constraint 2 takes the form of a configurable maximum depth of the lambdas, which is imposed to prevent our programs from growing too large.
However, as these lambdas can be nested, this hyperparameter alone is not enough to properly constrain the size of a program.
For instance, take a lambda as simple as \code{$\lambda$x $\to$ map otherLambda x}.
Depending on the allowed types, \code{otherLambda = $\lambda$x $\to$ map yetAnotherLambda x} would be a valid function and so on, which could lead to lambdas of arbitrarily large size.
Therefore, to prevent excessively large lambda nesting, we constrain nested functions to always be \code{$\lambda x \to x$} (the identity function).

To enforce these constraints, similarly to STGP, \algoname employs type-possibility tables to generate lambdas.  
For the main program tree, as the argument and output types are known beforehand, both STGP and \algoname only need to create two tables: one for the \emph{grow} and one for the \emph{full} method.
However, \algoname needs to generate lambdas involving every possible type allowed by the current program.  
Due to the recursive nature of the table, different argument types can lead up to vastly different type-possibility tables, so we need to keep one separate table for each possible argument type. 
As a corollary of Constraint 1, those tables are also guaranteed never to grow too large, as they never need to calculate possibilities for depths larger than the maximum lambda depth.
These tables also differ from the main tables in the sense that they only consider a node valid if at least one of its subtrees can have an argument leaf as a descendent, enforcing Constraint 2.

In terms of mutation and crossover, lambdas are treated as regular
terminals.  They are always discarded and regenerated (using the
process described above) or moved in their entirety, being treated
essentially as a single unit.

\subsection{Code Refinements}
\label{sec:code_refinements}
A well-known difficulty faced by GP algorithms is the occurrence of \emph{bloat}~\cite{luke2006comparison}, an unnecessary and uncontrollable growth of a program without any benefit to the fitness function. 
This happens naturally as some building blocks that apparently do not affect the program's output survive during successive applications of crossover and mutation. 
Not only do these bloats make the generated program longer and unreadable, but they can also affect the performance on the test set.
For example, consider the task of doubling a number and the candidate solution \code{x0 * (min x0 900)}. 
If the training set does not contain input cases such that \code{x0} $> 900$, then this will be a correct solution from the algorithm's point-of-view.

\citet{helmuth2017improving} empirically show that simpler programs often have a higher generalization capability, in addition to being easier to understand and reason about.
\citet{pantridge2022functional}, for example, applies a refinement step at the end of the search, repeatedly trying to remove random sections of the program and checking for improvements.

To alleviate the effect of bloats, we also apply a refinement procedure on the best tree found, considering the training data.
Refinement starts by applying simplification rules, which remove redundancies from the code:
\begin{itemize}
    \item Constant evaluations: if there are no argument terminals involved in a certain tree-branch, it can always safely be evaluated to a constant value, \emph{e.g.} \code{head [4*5, 1+2]} $\to$ \code{20};
    \item General law-application: the simplifier has access to a table of hand-written simplification procedures, which are known to be true (laws) (\emph{e.g.} \code{if True then a else b $\equiv$ a}, \code{a > a $\equiv$ False}, \code {length (singleton b) $\equiv$ 1}, etc).
\end{itemize}

\SetKwFunction{LocalSearch}{localSearch}%
\SetKwProg{Fn}{\textbf{function}}{\string:}{}%

\SetKwFunction{subtrees}{getMatchingSubtrees}%
\SetKwFunction{replace}{replace}%
\SetKwFunction{accuracy}{accuracy}%
\SetKwFunction{numberOfNodes}{nNodes}%
\SetKwFunction{getNext}{nextPreOrder}%
\SetKwFunction{hasNext}{hasNext}%
\SetKw{Continue}{continue}%

\begin{algorithm}[t]
\caption{The local search procedure}\label{alg:ls}
\DontPrintSemicolon

\Fn{\LocalSearch{tree}}{
    \lIf{not \hasNext{tree.node}}{\Return tree \label{lin:stop_crit}}
    \DataSty{bestTree} $\gets$ \DataSty{tree}\;
    \ForEach{$child \in \mathit{tree.node.children}$}{
        \If{child.outputType = tree.node.outputType}{
        \DataSty{newTree} $\gets$ \replace{tree.node, child}\;\label{lin:replace}
        \If{
        \accuracy{newTree} $\geq$ \accuracy{bestTree} \\
        \&\&
        \numberOfNodes{newTree} < \numberOfNodes{bestTree} \\
        }{
            \DataSty{bestTree} $\gets$ \DataSty{newTree}\;
        } 
        }
    }

    \If{bestTree $\neq$ tree}{
        \Return \LocalSearch{bestTree}\;
    }
    \Return \LocalSearch{\getNext{tree}}\;
    
}
\end{algorithm}

After this step, \algoname applies a Local Search procedure aiming at the removal of parts of the tree that do not contribute to, or even reduce, the correctness of the program considering the training set. 
The local search replaces each node with each one of its children and keeps the modified version if it improves or returns the same result.
Algorithm~\ref{alg:ls} describes this process. It takes as input \ArgSty{tree}, which has an internal representation of the current position being checked, that can be accessed via $tree.node$.
The procedure starts by calling \LocalSearch with the tree we obtained from the simplification rules, and the current position set to the tree root.
Next, the algorithm scans the children of the current node that have the same output type as their parent, and creates a new tree by replacing the parent node (Line~\ref{lin:replace}).
The best tree is stored, and the process continues recursively, advancing the current position to the next node in pre-order traversal if the tree is not changed, otherwise it will continue using the current position.
The process stops when there are no more positions to be checked (Line~\ref{lin:stop_crit}).

%% file: tables/functions_smaller.tex
\begin{table}[t!]
    \centering
    \caption{Functions supported by HOTGP.}
    \label{tab:functions}
    \begin{tabular}{l>{\raggedright\arraybackslash}p{4cm}}
    \toprule
    \textbf{Function Type} & \textbf{Function names}\\
    \midrule

    \code{Int} $\to$ \code{Int} $\to$ \code{Int} &
        \code{AddInt},
        \code{SubInt},
        \code{MultInt},
        \code{DivInt},
        \code{ModInt},
        \code{MaxInt},
        \code{MinInt}\\

    \code{Bool} $\to$ \code{Bool} & \code{Not}\\

    \code{Bool} $\to$ \code{Bool} $\to$ \code{Bool} &
        \code{And},
        \code{Or}\\

    \code{Bool} $\to$ \code{a} $\to$ \code{a} $\to$ \code{a} & \code{If}\\

    \code{Float} $\to$ \code{Float} & \code{Sqrt}  \\

    \code{Float} $\to$ \code{Float} $\to$ \code{Float} &
        \code{AddFloat},
        \code{SubFloat},
        \code{MultFloat},
        \code{DivFloat}\\

    \code{a} $\to$ \code{[a]} & \code{Singleton} \\
    \code{a} $\to$ \code{[a]} $\to$ \code{[a]} & \code{Cons} \\
    \code{[a]} $\to$ \code{a} & \code{Head} \\
    \code{[a]} $\to$ \code{[a]} & \code{Reverse} \\
    \code{[[a]]} $\to$ \code{[a]} & \code{Concat} \\

    \code{a} $\to$ \code{b} $\to$ \code{(a,b)} & \code{ToPair} \\
    \code{(a,b)} $\to$ \code{a} & \code{Fst} \\
    \code{(a,b)} $\to$ \code{b} & \code{Snd} \\

    \code{Char} $\to$ \code{Char} $\to$ \code{Bool} & \code{EqChar} \\
    \code{Char} $\to$ \code{Bool} &
        \code{IsLetter},
        \code{IsDigit} \\

    \code{Int} $\to$ \code{Float} & \code{IntToFloat} \\
    \code{Float} $\to$ \code{Int} & \code{Floor} \\

    \code{Int} $\to$ \code{Int} $\to$ \code{Bool} &
        \code{GtInt},
        \code{LtInt},
        \code{EqInt} \\

    \code{[a]} $\to$ \code{Int} & \code{Len} \\
    \code{Int} $\to$ \code{[a]} $\to$ \code{[a]} & \code{Take} \\
    \code{Int} $\to$ \code{Int} $\to$ \code{Int} $\to$ \code{[Int]} & \code{Range} \\
    \code{[Int]} $\to$ \code{Int} &
        \code{SumInts},
        \code{ProductInts} \\

    \code{[Float]} $\to$ \code{Float} &
        \code{SumFloats},
        \code{ProductFloats} \\

    \code{[[Char]]} $\to$ \code{[Char]} & \code{Unlines} \\
    \code{Int} $\to$ \code{[Char]} & \code{ShowInt} \\
    \code{[a]} $\to$ \code{[b]} $\to$ \code{[(a,b)]} & \code{Zip} \\
    \code{(a} $\to$ \code{b)} $\to$ \code{[a]} $\to$ \code{[b]} & \code{Map} \\
    \code{(a} $\to$ \code{Bool)} $\to$ \code{[a]} $\to$ \code{[a]} & \code{Filter} \\

    \bottomrule
    \end{tabular}
\end{table}

%% file: sections/results.tex
\vspace{1em}
In this section, we compare \algoname to state-of-the-art GP-based program synthesis algorithms found in the literature. 
For this comparison, we employ the ``General Program Synthesis Benchmark Suite''~\cite{helmuth2015detailed}, which contains a total of $29$ benchmark problems for inductive program synthesis\footnote{The full source code for \algoname can be downloaded from: \href{https://github.com/mcf1110/hotgp}{https://github.com/mcf1110/hotgp}.}.

Following the recommended instructions provided by \citet{helmuth2015detailed}, we executed the algorithm using $100$ different seeds for each benchmark problem. 
We used the recommended number of training and test instances and included the fixed edge cases in the training data.
We also used the same fitness functions described in their paper.

For the evolutionary search, we used a steady-state replacement of $2$ individuals per step, with an initial population of $1\,000$, and using a Parent-Scalar of $99.93\%$.
The maximum tree depth was set to $15$ for the main program and $3$ for $\lambda$-functions.
The crossover and mutation rates were both empirically set to $50\%$. 
We allowed a maximum of $300\,000$ evaluations with an early stop whenever the algorithm finds a perfectly accurate solution according to the training data.

We report the percentage of correct solutions found within the $100$ executions taking into consideration the training and test data sets, before and after the refinement process. 
To position such results with the current literature, we compare the obtained results against those obtained by PushGP~\cite{helmuth2015detailed}, Grammar-Guided Genetic Programming (G3P)~\cite{g3p}, and the extended grammar version of G3P (here called G3P+)~\cite{g3pe}, and some recently proposed methods such as Code Building Genetic Programming (CBGP)~\cite{pantridge2022functional}, and G3P with Haskell and Python grammars (G3Phs and G3Ppy)~\cite{garrow2022functional}. 
We have not compared with STGP and PolyGP since their original papers~\cite{montana1995strongly,yu2001polymorphism} predate this benchmark suite. 
All the obtained results are reported in Table~\ref{tab:comparison}. 
In this table, all the benchmarks that could not be solved with our current function set are marked with ``--'' in \algoname columns. 
For the other approaches, the dash mark means the authors did not test their algorithm for that specific benchmark.

\subsection{Analysis of the results}
\enlargethispage{-2em}

\input{tables/comparison}

Compared to the other algorithms, \algoname has the highest success rate for the test set in $9$ of the benchmark problems, followed by PushGP and CBGP, which got the highest rate for $7$ and $5$ of the benchmarks, respectively.
An important point to highlight is that \algoname obtained a $100\%$ success rate in $4$ problems, and a $\geq 75\%$ in $7$, a result only matched by CBGP.
Moreover, \algoname obtained at least a $50\%$ success rate in $10$ out of the 29 problems, which is not matched by any of the compared methods.
This brings evidence to our initial hypothesis that including type information in the program synthesis can, indeed, reduce the search space to improve the efficiency of the evolution process.

For example, in the \emph{compare-string-lengths} problem, the input arguments are of the type \code{String}, and the output is a \code{Bool} but allowing intermediate \code{Int} type. 
Looking at Table~\ref{tab:functions}, we can see that there are a few ways to convert a string to a boolean, as we only support functions in the character level. 
The best we can do is to extract the first character with \code{Head} and then convert the character into a boolean with \code{IsLetter} or \code{IsDigit}. 
We could, for instance, generate a program that does that for both inputs and compares the results with different boolean operators. 
We could also apply a \code{Map} function before applying \code{Head}. 
Also, to convert a string into an integer, the only solution is to use the \code{Length} function and the few combinations on how to convert two integers into a boolean. 
One example of obtained solution is \code{((length x1) > (length x0)) \&\& ((length x1) < (length x2))}.

On the other hand, for the \emph{last-index-of-zero} problem, a possible correct solution using our grammar is \code{fst (head (reverse (filter ($\lambda$y $\to$ 0 == (snd y)) (zip (range 0 (length x0) 1) x0))))}. 
So the synthesizer must first enumerate the input, apply a filter to keep only the elements that contain $0$, reverse the list, take the first element, and return its index. 
One of the best obtained solutions was \code{((length x0) + (if ((head (reverse x0)) == 0) then 1 else 0)) - 2} with $32\%$ of accuracy. 
It simply checks if the last element is $0$, if it is, it returns the length of the list minus one, otherwise it returns the length minus two. 
This is a possible general case for a recursive solution where it checks the last element and, if it is not zero, recurses with the remainder of the list.

As described in Section~\ref{sec:code_refinements}, the code refinement step always produces an equal or better solution. 
These improvements are more noticeable on the \emph{median} and \emph{for-loop-index} problems.
This is due to the fact that code refinement is sometimes capable of discarding misused numerical constants. 
For example, one solution to the \emph{median} problem with $99\%$ of accuracy on the training set was \code{max -96 (min (max x1 x2) (max (min x1 x2) x0))} 
that only works if the median of the three arguments is greater than $-96$, otherwise it will always return a constant value. 
After the code refinements, \algoname finds the final and correct solution:
\code{min (max (min x2 x1) x0) (max x1 x2)}.

\begin{figure*}[ht]
    \centering
    \includegraphics[width=\linewidth]{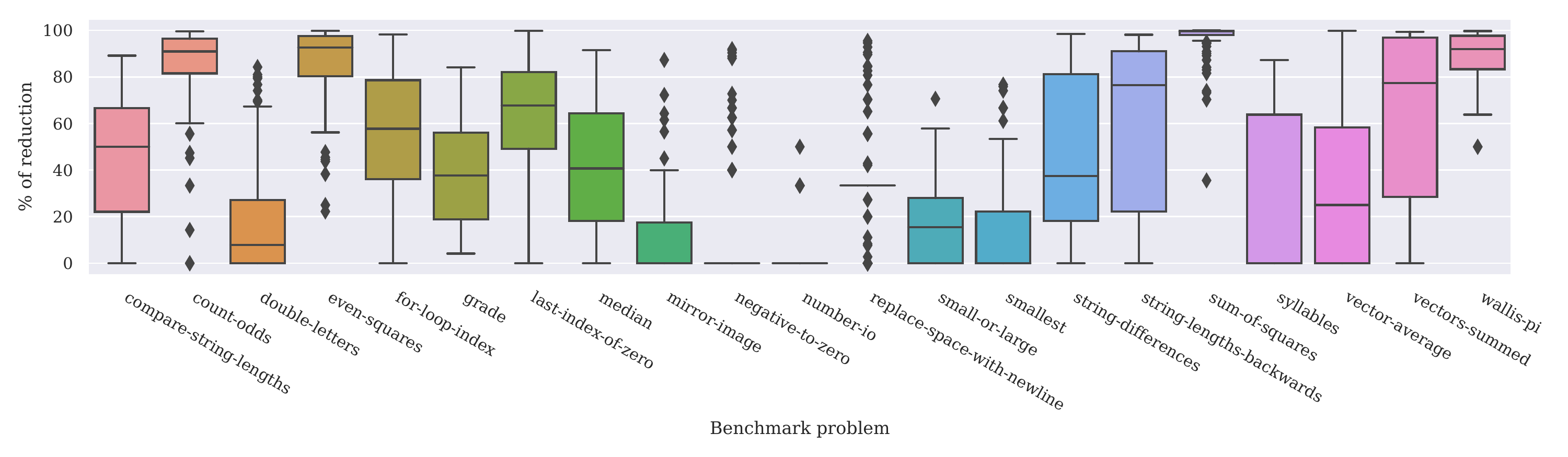}
    \vspace{-2em}
    \caption{Percentage of reduction in the number of nodes caused by the refinement process.}
    \label{fig:refinement}
\end{figure*}

Another benefit of code refinement is reducing the program size, which can improve the readability of the generated program. Figure~\ref{fig:refinement} shows the rate of decrease in the program size after refinements, with a geometric mean of $52\%$.
The refinement process effectiveness varies depending on the nature of the solutions of the problem.
For most problems, the end of the upper quartile is well within the $>75\%$ reduction mark, meaning it was not unusual for some solutions to get largely simplified.
However, more evident results are yielded in problems such as \emph{counts-odds}, \emph{even-squares}, and \emph{sum-of-squares}, which dealt with fewer types (and thus a reduced grammar) and usually reached the maximum evaluation count, therefore were more susceptible to bloat. 
Notably, \emph{number-io} and \emph{negative-to-zero} had almost no reduction, showing the algorithm could directly find a perfect and near minimal solution.

To provide further insights into how minimal the correct solutions actually are, and how susceptible to bloat each problem is, \autoref{tab:compare_to_manual} takes the smallest correct solution that \algoname could find for each problem, and compares them to the handwritten solutions crafted by the authors. Even before the refinement procedure, most of the solutions have the same node count as a the handwritten ones, and nearly all of them are reasonably close. The \emph{sum-of-squares} was initially much larger than the manual solution, but after refinement the size reduction is notable.
The only correct solution we found for \emph{mirror-image} also has the biggest reduction of the batch, showing a $87\%$ reduction overall.

\begin{table}[ht]
    \small
    \centering
    \caption{Node count of the hand-crafted solutions (HC) and the smallest correct solutions found by \algoname and HOTGP*. The node count relative to HC is shown in parenthesis.}
    \vspace{-1em}
    \label{tab:compare_to_manual}
    \begin{tabular}{lrrr}
    \toprule
    Benchmark & HC & \algoname & \algoname* \\
    \midrule
    compare-string-lengths     &    11 &     11  (1.0x) &      11 (1.0x) \\
    count-odds                 &     4 &      4  (1.0x) &       4 (1.0x) \\
    for-loop-index             &     7 &     25  (3.6x) &       9 (1.3x) \\
    grade                      &    27 &     45  (1.7x) &      29 (1.1x) \\
    median                     &     9 &      9  (1.0x) &       9 (1.0x) \\
    mirror-image               &    10 &    102 (10.2x) &      13 (1.3x) \\
    negative-to-zero           &     3 &      3  (1.0x) &       3 (1.0x) \\
    number-io                  &     4 &      4  (1.0x) &       4 (1.0x) \\
    replace-space-with-newline &     8 &      8  (1.0x) &       8 (1.0x) \\
    small-or-large             &    11 &     12  (1.1x) &      12 (1.1x) \\
    smallest                   &     7 &      7  (1.0x) &       7 (1.0x) \\
    string-lengths-backwards   &     4 &      5  (1.2x) &       5 (1.2x) \\
    sum-of-squares             &     7 &    163 (23.3x) &      30 (4.3x) \\
    vector-average             &     6 &      6  (1.0x) &       6 (1.0x) \\
    vectors-summed             &     5 &      5  (1.0x) &       5 (1.0x) \\
    \bottomrule
    \end{tabular}
    \vspace{-1em}
\end{table}

%% file: tables/comparison.tex

\begin{table*}[t]
\centering
\caption{Successful solutions found for each problem (\% of executions) considering the training (Tr) and test (Te) data sets. \algoname* lists the results obtained with \algoname after the simplification procedure. The best values for the test data sets of each problem are highlighted. The 
\emph{checksum}, \emph{collatz-numbers}, \emph{string-differences}, \emph{wallis-pi} and \emph{word-stats} problems are ommitted as no algorithm was able to find results for those problems.
}
\label{tab:comparison}
\begin{tabular}{l|rr|rr|r|r|rr|r|rr|rr}
\toprule
 & \multicolumn{2}{c|}{\algoname} & \multicolumn{2}{c|}{\textbf{\algoname*}} & PushGP &  G3P & \multicolumn{2}{c|}{G3P+} & CBGP & \multicolumn{2}{c|}{G3Phs} & \multicolumn{2}{c}{G3Ppy} \\
Benchmark &  Tr &  Te &  Tr &  Te &  Te &  Te &  Tr & Te &  Te &  Tr &  Te &  Tr &  Te \\
\midrule

compare-string-lengths & 100 & \underline{100} & 100 & \underline{100} & 7 & 2 & 96 & 0 & 22 & 94 & 5 & 12 & 0 \\
count-odds & 46 & 46 & 50 & \underline{50} & 8 & 12 & 4 & 3 & 0 & -- & -- & -- & -- \\
digits & -- & -- & -- & -- & \underline{7} & 0 & 0 & 0 & 0 & -- & -- & -- & -- \\
double-letters & 0 & 0 & 0 & 0 & \underline{6} & 0 & 0 & 0 & -- & -- & -- & -- & -- \\
even-squares & 0 & 0 & 0 & 0 & \underline{2} & 1 & 0 & 0 & -- & -- & -- & -- & -- \\
for-loop-index & 73 & 39 & 73 & \underline{59} & 1 & 8 & 9 & 6 & 0 & -- & -- & -- & -- \\
grade & 37 & 32 & 39 & \underline{37} & 4 & 31 & 63 & 31 & -- & -- & -- & -- & -- \\
last-index-of-zero & 0 & 0 & 0 & 0 & 21 & 22 & 97 & \underline{44} & 10 & 0 & 0 & 2 & 2 \\
median & 82 & 73 & 100 & \underline{99} & 45 & 79 & 99 & 59 & 98 & 100 & 96 & 39 & 21 \\
mirror-image & 1 & 1 & 1 & 1 & 78 & 0 & 89 & 25 & \underline{100} & -- & -- & -- & -- \\
negative-to-zero & 100 & \underline{100} & 100 & \underline{100} & 45 & 63 & 24 & 13 & 99 & 0 & 0 & 68 & 66 \\
number-io & 100 & \underline{100} & 100 & \underline{100} & 98 & 94 & 95 & 83 & \underline{100} & 100 & 99 & 100 & \underline{100} \\
pig-latin & -- & -- & -- & -- & 0 & 0 & 4 & \underline{3} & -- & -- & -- & -- & -- \\
replace-space-with-newline & 38 & 38 & 38 & 38 & \underline{51} & 0 & 29 & 16 & 0 & -- & -- & -- & -- \\
scrabble-score & -- & -- & -- & -- & \underline{2} & \underline{2} & 1 & 1 & -- & -- & -- & -- & -- \\
small-or-large & 28 & \underline{59} & 28 & \underline{59} & 5 & 7 & 39 & 9 & 0 & 30 & 4 & 0 & 0 \\
smallest & 98 & 95 & 100 & \underline{100} & 81 & 94 & 100 & 73 & \underline{100} & 100 & \underline{100} & 99 & 89 \\
string-lengths-backwards & 87 & 87 & 89 & \underline{89} & 66 & 68 & 20 & 18 & -- & 0 & 0 & 35 & 34 \\
sum-of-squares & 1 & 1 & 1 & 1 & \underline{6} & 3 & 5 & 5 & -- & -- & -- & -- & -- \\
super-anagrams & -- & -- & -- & -- & 0 & 21 & 43 & 0 & -- & 30 & 5 & 51 & \underline{38} \\
syllables & 0 & 0 & 0 & 0 & 18 & 0 & 53 & \underline{39} & -- & -- & -- & -- & -- \\
vector-average & 78 & 79 & 80 & 80 & 16 & 5 & 0 & 0 & \underline{88} & 67 & 4 & 0 & 0 \\
vectors-summed & 34 & 34 & 37 & 37 & 1 & 91 & 28 & 21 & \underline{100} & 100 & 68 & 0 & 0 \\
x-word-lines & -- & -- & -- & -- & \underline{8} & 0 & 0 & 0 & -- & -- & -- & -- & -- \\


\midrule
\textbf{\# of Best Results} &  & 4 &  & \underline{9} & 7 & 2 & & 3 & 5 & & 1 &  & 2  \\
\midrule
$\mathbf{= 100\%}$   &  & 3 &  & \underline{4}  & 0 & 0 &  & 0 & \underline{4} & & 1 &  & 1 \\
$\mathbf{\geq 75\%}$ &  & 6 &  & \underline{7}  & 3 & 4 &  & 1 & \underline{7} & & 3 &  & 2 \\
$\mathbf{\geq 50\%}$ &  & 8 &  & \underline{10} & 5 & 6 &  & 3 & 7             & & 4 &  & 3 \\

\bottomrule
\end{tabular}
\end{table*}
\vspace{4em}

%% file: sections/conclusion.tex
\balance
This paper presents \algoname, a GP algorithm that supports higher-order functions, $\lambda$-functions, polymorphic types, and the use of type information to constrain the search space. It also sports a grammar based on the Haskell language using only pure functions in the nonterminals set.
Our main arguments in favor of this approach are: i) limiting our programs to pure functions avoids undesirable behaviors; ii) using type-level information and parametric polymorphism reduces the search space directing the GP algorithm towards the correct solution; iii) higher-order functions eliminate the need of several imperative-style constructs (\emph{e.g.}, for loops).

\algoname differs from most GP implementations as it actively uses the information of input and output types to constrain the candidate terminals and nonterminals while creating new solutions or modifying existing ones, and to select feasible points of recombination.

We have evaluated our approach with $29$ benchmark problems and compared the results with $6$ state-of-the-art algorithms from the literature. 
Overall, we got favorable results, consistently returning a correct program most of the time for $10$ problems, a mark that was not met by any of the tested methods.
Moreover, \algoname achieved the highest success rates more often than the state of art. 

We also applied code refinements to the best solution found by the algorithm to reduce the occurrence of \emph{bloat} code. This procedure leads to further improvements in the results while at the same time improving the readability of the final program.

Even though we achieved competitive results, we observed that there are still possible improvements. 
First, our nonterminals set is much smaller than some of the state-of-the-art algorithms (\emph{e.g.}, PushGP). 
Future work includes carefully examining the impact of adding new functions to the grammar. 
This inclusion might further simplify the PS or allow us find solutions that are not currently being found. 
On the other hand, including new functions also expands the search space and can hinder some of our current results.

Our approach could also benefit from a more modular perspective for PS. 
In a modular approach, the problem is first divided into simpler tasks which are solved independently and then combined to create the complete synthesized program. 
This approach will require support to different forms of functional composition and the modification of the benchmark to create training data for the different subtasks. 
Such a synthesizer could also be coupled with Wingman\footnote{\href{https://haskellwingman.dev/}{https://haskellwingman.dev/}} (the current implementation of advanced Haskell code generation), which can either synthesize the whole program or guide the process using only the type information, and code holes.

Further research is also warranted concerning more advanced type-level information such as Generalized Algebraic Data Types (GADTs), Type Families, Refinement Types and Dependent Types.
More type information could further constrain the search space and, in some situations, provide additional hints to the synthesis of the correct program. 
Clearly, this must be accompanied by a modification of the current benchmarks and the inclusion of new benchmarks that provides this high-level information about the desired program.